\documentclass{article}

\usepackage[main, final]{neurips_2026}

\usepackage[utf8]{inputenc} 
\usepackage[T1]{fontenc}    
\usepackage{hyperref}       
\usepackage{url}            
\usepackage{booktabs}       
\usepackage{amsfonts}       
\usepackage{nicefrac}       
\usepackage{microtype}      
\usepackage{xcolor}         

\usepackage{xspace}
\usepackage{amsmath}
\usepackage{bbm}
\usepackage{graphicx}
\usepackage{multirow}
\usepackage{subcaption}
\usepackage{wrapfig}
\usepackage{listings}
\usepackage{xcolor}
\usepackage{float}

\lstdefinestyle{prompt}{
  basicstyle=\small\ttfamily,
  breaklines=true,
  breakatwhitespace=true,
  breakautoindent=false,
  columns=fullflexible,
  backgroundcolor=\color{gray!5},
  frame=l,
  framerule=1.5pt,
  rulecolor=\color{gray!40},
  xleftmargin=10pt,
  aboveskip=10pt,
  belowskip=10pt,
  literate={--}{\textendash}{1}
}

\AtBeginDocument{%
}

\title{FreeAct: Demonstration-Free Robot Adaptation via Action-Grounded Generated Videos}

\author{%
  Zhe-Han Mo \quad Jia-Ning Li \quad Jiaxi Song \quad
  Meng-Hao Guo\textsuperscript{$\dagger$} \quad
  Shi-Min Hu\thanks{Corresponding author.} \\
  BNRist, Department of Computer Science and Technology, Tsinghua University, China\\
  \texttt{\{mozh23,ljn24,songjx22\}@mails.tsinghua.edu.cn} \\
  \texttt{\{gmh,shimin\}@tsinghua.edu.cn}.
}

\makeatletter
\DeclareRobustCommand\onedot{\futurelet\@let@token\@onedot}
\def\@onedot{\ifx\@let@token.\else.\null\fi\xspace}
\makeatother

\begin{document}

\maketitle

\begingroup
\renewcommand{\thefootnote}{\fnsymbol{footnote}}
\footnotetext[2]{Project lead.}
\endgroup

\begin{abstract}
  Adapting pretrained Vision-Language-Action (VLA) policies to new deployment environments typically requires collecting expert demonstrations in each target domain, which is costly and difficult to scale.
  In this paper,
  we present \textbf{FreeAct}, a framework for \emph{demonstration-free robot adaptation} that converts generated videos into physically grounded supervision.
  A key challenge is that generated manipulation videos, while visually plausible, often contain embodiment artifacts and kinematic inconsistencies that make direct inverse-dynamics labeling unreliable.
  FreeAct addresses this challenge by learning a discrete latent action space from large-scale multi-lab robot videos and aligning it with camera-frame end-effector SE(3) motion and gripper-state changes. The resulting latent action tokenizer captures task-relevant motion while reducing reliance on embodiment-specific visual artifacts, enabling generated target-domain videos to be pseudo-labeled with reliable latent action tokens. These tokens are first injected into VLA pretraining and then used, together with source-domain action supervision, to adapt the policy to unseen environments.
  Empirically, FreeAct improves success from 2\% to 57\% under target-domain shift using generated videos alone, and further to 74\% through self-evolution guided by a multimodal large language model (MLLM) on the policy's own rollouts, without expert demonstrations.
  Further analysis shows that the learned latent action space forms a
  kinematically structured and transferable representation,
  establishing generated videos as a scalable source of supervision for robot manipulation. Code and data are available at \url{https://freeact.pages.dev}.
\end{abstract}

\section{Introduction}

Building robots that can follow open-ended instructions and operate reliably across diverse, unseen environments is a long-standing goal in robot learning~\citep{Brohan23RT-1, Zitkovich23RT-2, Guo26RoboDawn, Guo26nanoAstra}.
VLA models have made substantial progress toward this goal: pretrained on large-scale robot datasets and tuned on target-environment demonstrations, a single policy can perform diverse manipulation tasks across objects, scenes, and embodiments~\citep{Black25pi05, Gemini25Robotics15, NVIDIA25GR00TN1}. 
Yet the last step, tuning on target-environment demonstrations, remains a key bottleneck for deployment. Policies trained in one environment overfit to its distribution and degrade sharply under changes in lighting, texture, camera viewpoint, distractors, object instances, or robot appearance~\citep{Yu23ROSIE, Fei25LIBERO-Plus}, while collecting demonstrations for every new environment is costly, slow, and difficult to scale~\citep{ONeill24OXE}.
This motivates a practically important setting we term \emph{demonstration-free robot adaptation}: adapting a pretrained policy to a new environment without any target-environment expert demonstrations.

Recent advances in video generation~\citep{Zheng26T2VFire} offer a compelling path toward this goal. Trained on internet-scale corpora, foundation video generators can synthesize visually realistic manipulation videos in a zero-shot manner from language instructions and reference images, naturally covering the broad range of appearance variation, scene diversity, and long-tail conditions~\citep{Brooks24Sora, NVIDIA25Cosmos}.
In principle, such generated videos could serve as deployment-specific experience for adapting pretrained VLA policies, eliminating the need for target-environment teleoperation. The central challenge is recovering executable robot actions from these videos. A natural approach is to synthesize manipulation videos and then apply an Inverse Dynamics Model (IDM)~\citep{Du23UniPi, Tian25Seer}.
However, we empirically find this pipeline to be unreliable when the video generator is used without domain-specific finetuning. Generated videos look plausible, but often contain kinematic inconsistencies such as non-rigid arm deformations. These artifacts fall outside the IDM's training distribution and corrupt the inferred actions~\citep{Liu26Geometry}. 
Tuning the generator on robot-specific data could suppress these artifacts, 
but is computationally expensive and risks eroding its broad visual coverage and physical priors~\citep{Jang25DreamGen}. We therefore ask: \emph{can generated videos be converted into physically grounded action supervision for robot adaptation, without finetuning the generator?}

To this end, we propose \textbf{FreeAct}, a three-stage framework that learns physically grounded latent action representations and uses them to adapt pretrained VLA policies from generated videos. In the first stage, we train a latent action model (LAM) on large-scale robot videos with a VQ-VAE objective~\citep{van17VQ-VAE}, augmented by an auxiliary Action Decoder that supervises the latent bottleneck to predict camera-frame end-effector motion. This grounding pressures the latent space to encode kinematics rather than embodiment appearance, yielding a representation that remains reliable on generated videos. In the second stage, the resulting discrete tokens serve as prediction targets for VLA pretraining, injecting a physically grounded motion abstraction into the policy backbone without requiring action labels. In the third stage, we attach a continuous action head and jointly finetune on action-labeled source data and generated target-environment videos pseudo-labeled by the stage-1 model, with a projected gradient update to reconcile the two data streams. Together, this pipeline converts generated videos into action-grounded supervision for demonstration-free policy adaptation.

The practical value of FreeAct lies in its ability to close the full
adaptation loop without expert intervention. Generated target-environment
videos alone lift a pretrained VLA from $2\%$ to $57\%$ success under
substantial domain shift. The adapted policy then refines itself: a
handful of its own rollouts, selected by a multimodal LLM judge,
raise success to $74\%$. The two phases are complementary: generated
videos bootstrap broad task competence, while self-refinement corrects
residual errors such as depth ambiguities that generated supervision
alone cannot resolve.
A suite of diagnostics reveals a consistent pattern: reconstruction-only baselines yield fragmented, appearance-driven latent spaces, whereas our physically grounded model induces a continuous action manifold whose neighborhoods correspond to similar end-effector motions. This structural alignment is what enables FreeAct to turn generated videos from visually plausible trajectories into actionable supervision for robot adaptation.

Our contributions are threefold:
\begin{enumerate}
    \item We propose FreeAct, a three-stage framework that
    grounds discrete latent actions in camera-frame end-effector
    kinematics, transfers this motion abstraction into a pretrained
    VLA, and adapts the policy from action-grounded generated videos
    to a target environment without expert demonstrations.
    \item We introduce a set of diagnostics for evaluating latent
    action spaces and show that physical grounding produces a
    continuous, linearly decodable kinematic manifold, in contrast to
    the fragmented, appearance-driven representations learned by
    reconstruction-only baselines.
    \item We show that FreeAct enables demonstration-free
    adaptation under aggressive domain shift, lifting a pretrained VLA
    from $2\%$ to $57\%$ success with generated videos alone, and to
    $74\%$ through self-evolution on a handful of the policy's own
    rollouts.
\end{enumerate}

\section{Related Work}

\paragraph{Vision-Language-Action models}
VLA models adapt pretrained MLLMs~\citep{Bai25Qwen3-VL, Zhang26Bee} to robot
control, predicting action tokens or continuous action chunks from
visual observations and language
instructions~\citep{Zitkovich23RT-2, Kim25OpenVLA,
Black25pi0, NVIDIA25GR00TN1}. Recent generalist policies scale this paradigm with
large robot datasets, cross-embodiment mixtures, expressive action
decoders, and richer conditioning signals such as memory, subtask
prompts, subgoal images, and embodiment-aware
prompts~\citep{ONeill24OXE, Ghosh24Octo, Zheng26X-VLA, Intelligence26pi07}.
Despite their growing generality,
deploying these policies to a new environment typically still
requires target-domain robot data, whether through demonstrations,
post-training, or carefully curated cross-embodiment pretraining. We
target the more practical regime in which no such target-domain
robot data is provided.

\paragraph{Video generation for robot learning}
Video generators provide spatio-temporal priors increasingly exploited for robot learning. One line executes generated videos at inference time, recovering actions with an IDM or a goal-conditioned policy~\citep{Du23UniPi, Du24VLP} or by tracking a tool or object across frames~\citep{Liang25Dreamitate, Patel26Robotic}; Gen2Act~\citep{Bharadhwaj25Gen2Act} instead conditions a policy on generated human videos while learning actions from robot demonstrations. Another line trains a world model, typically from a pretrained generator, and either labels or evaluates its rollouts~\citep{Jang25DreamGen, Guo26Ctrl-World, Gao26DreamDojo} or predicts actions jointly with future visual states~\citep{Zhu25UWM, Bi25Motus, Kim26CosmosPolicy, Li26LingBotVA, Ye26DreamZero, Lyu26LDA-1B}. We instead keep the generator off-the-shelf and convert its videos into action supervision through a physically grounded LAM.

\paragraph{Latent action learning}
Another line of work learns latent actions from action-free videos
by encoding inter-frame transitions through a discrete bottleneck
and using the resulting codes for control or policy
pretraining~\citep{Bruce24Genie, Schmidt24LAPO, Cui24DynaMo,
Ye25LAPA, Bu25UniVLA, Chen26villa-X}. Such objectives primarily
constrain visual predictability, leaving the relationship between
latent codes and executable robot motion only indirectly specified.
\citet{Liu25Videos} align video-derived latent actions with
environment actions through online interaction; we instead retain
the action-free latent objective and ground it offline with
end-effector motion from a calibrated subset of a multi-lab robot
corpus.


\section{Method}

\paragraph{Problem setup}
We study \emph{demonstration-free robot adaptation}: given a VLA policy pretrained
on source environments, our goal is to adapt it to a target environment without
any target-environment expert demonstrations.
Let
\[
  \mathcal{D}_{\mathrm{src}}
  = \{(o_t,\,o_{t+\Delta},\,\ell_t,\,a_t)\}
\]
denote action-labeled robot data from source environments, where $o_t$ and
$o_{t+\Delta}$ are RGB observations separated by a fixed temporal horizon
$\Delta$, $\ell_t$ is a language instruction, and
\[
  a_t = [\Delta p_t,\,\Delta r_t,\,\Delta g_t] \in \mathbb{R}^7
\]
is the camera-frame end-effector action comprising 3D translation
$\Delta p_t \in \mathbb{R}^3$, 3D axis-angle rotation
$\Delta r_t \in \mathbb{R}^3$, and scalar gripper-state change
$\Delta g_t \in \mathbb{R}$.
In addition, we have access to videos synthesized by an off-the-shelf video
generator for the target domain,
\[
  \mathcal{D}_{\mathrm{gen}}
  = \{(o_t^g,\,o_{t+\Delta}^g,\,\ell_t^g)\},
\]
which provide visual trajectories under target-domain appearance but carry no
real robot actions.
In the \emph{self-evolution} extension, a small number of the adapted
policy's own target-environment rollouts, selected by an automated
quality signal, form $\mathcal{D}_{\mathrm{self}}$.

FreeAct addresses this setting through three stages: Stage~1 learns a \emph{physically grounded LAM} from large-scale
multi-lab robot videos, using a calibrated subset to anchor the latent space to
camera-frame kinematics.
Stage~2 uses the resulting discrete latent action tokens as supervision for VLA
pretraining, injecting motion abstraction into the policy backbone without
requiring dense action labels.
Stage~3 adapts the pretrained VLA to the target environment by jointly
optimizing over source action-labeled data and pseudo-labeled generated videos.
At test time, the adapted policy maps a target-environment observation and
language instruction directly to a continuous robot action.

\subsection{Physically grounded latent action learning}
\label{sec:lam}

\begin{figure}[t]
  \centering
  \includegraphics[width=\textwidth]{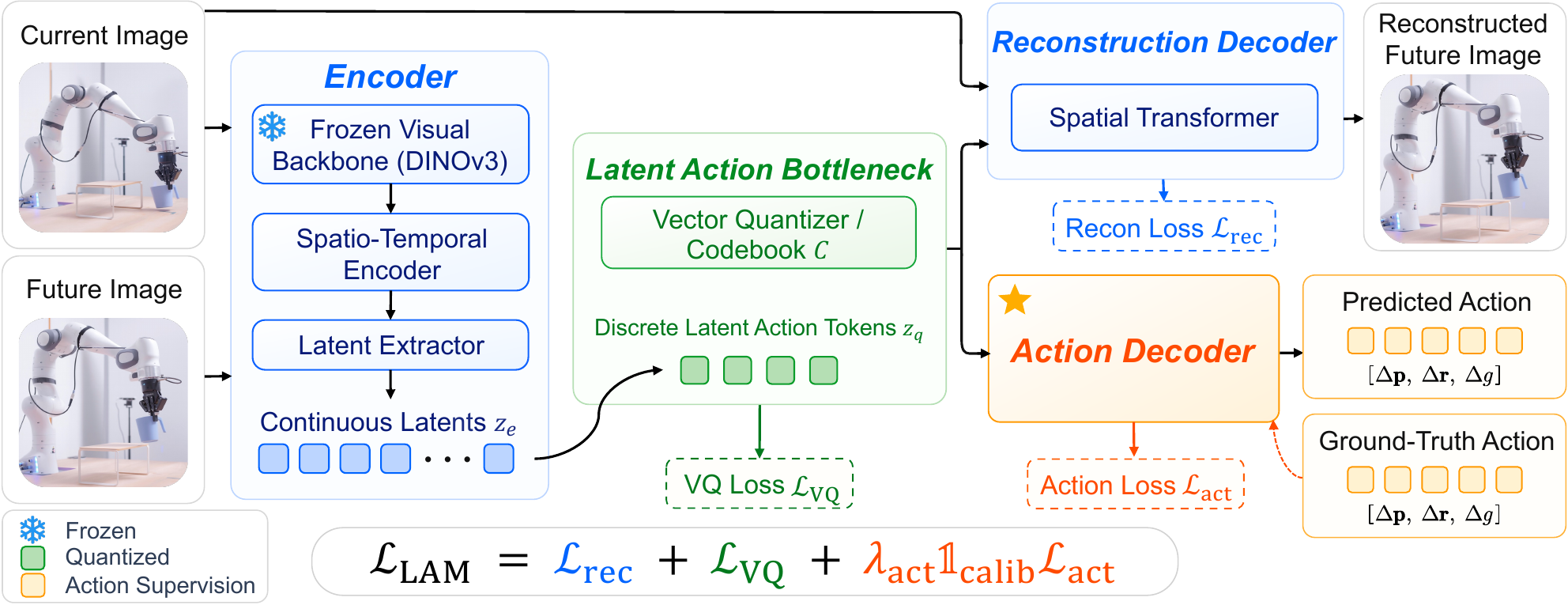}
  \caption{\textbf{Stage-1 latent action model.} A frozen DINOv3 encoder maps a frame pair to continuous latents $z_e$, vector-quantized into discrete tokens $z_q$ against codebook $\mathcal{C}$. A reconstruction decoder predicts the future frame ($\mathcal{L}_{\mathrm{rec}}$, $\mathcal{L}_{\mathrm{VQ}}$). Our key addition (\(\star\)) is an Action Decoder that maps $z_q$ to the camera-frame end-effector action and is supervised on the calibrated subset ($\mathcal{L}_{\mathrm{act}}$), grounding the codebook in executable kinematics.}
  \label{fig:lam_architecture}
\end{figure}

Stage~1 builds on the standard latent action
model~\citep{van17VQ-VAE, Schmidt24LAPO} to learn a discrete latent
representation of motion from frame pairs and ground it in executable
robot kinematics. A spatio-temporal encoder $Q_\phi$ maps a frame pair
$(o_t, o_{t+\Delta})$ into a sequence of $K$ continuous latents, which
are vector-quantized against a learnable codebook $\mathcal{C}$ to
yield discrete tokens
\[
  z_t = [z_t^{(1)}, \dots, z_t^{(K)}],
  \qquad z_t^{(k)} \in \{1,\dots,|\mathcal{C}|\},
\]
and a decoder $D_{\mathrm{rec}}$ reconstructs the future frame from
$z_t$ and $o_t$. We use a frozen DINOv3 ViT~\citep{Simeoni25DINOv3}
as the visual backbone and a transformer encoder--decoder~\citep{Xu22Transformers,Bao25MultiwayT2I} for the
spatio-temporal and reconstruction modules; full architectural
specifications are deferred to~\autoref{app:hyperparams}. \autoref{fig:lam_architecture} illustrates the resulting pipeline, including our auxiliary Action Decoder branch.

\paragraph{Physical grounding}
Trained with reconstruction alone, this template offers no pressure
on \emph{what} the latent space encodes. The bottleneck is free to organize itself around appearance, texture, or embodiment-specific cues, and prior LAMs exhibit exactly this failure: fragmented, appearance-driven codebooks whose neighborhoods carry little kinematic meaning (\autoref{sec:exp_rq1}).
We address this with an auxiliary Action Decoder
\[
  \hat{a}_t = D_{\mathrm{act}}(z_t),
\]
a lightweight multi-layer perceptron (MLP) that maps the quantized latents directly to a
predicted camera-frame end-effector action. The branch is supervised
only on the calibrated subset of $\mathcal{D}_{\mathrm{src}}$ for
which camera--robot extrinsics are available.

The design decision is to use the \emph{camera frame} rather than a
shared world frame as the supervision target. Camera extrinsics vary
substantially across laboratories and episodes, so a model
that shortcuts action prediction via embodiment appearance or
scene layout cannot generalize across episodes. 
The latent space is instead pressured to encode end-effector motion
relative to the camera in a manner that generalizes across viewpoints. This inductive bias remains valid on generated videos: while full-arm geometry is
frequently hallucinated, end-effector motion relative to the scene
tends to stay faithful to the task.

\paragraph{Training objective}

The stage-1 loss combines reconstruction, VQ commitment, and the
action supervision above:
\begin{equation}
  \mathcal{L}_{\mathrm{LAM}}
  = \mathcal{L}_{\mathrm{rec}}
  + \mathcal{L}_{\mathrm{VQ}}
  + \lambda_{\mathrm{act}}\,
    \mathbbm{1}_{\mathrm{calib}}\,
    \mathcal{L}_{\mathrm{act}}
  \label{eq:lam_total}
\end{equation}
where $\mathbbm{1}_{\mathrm{calib}}$ gates the action loss to the
calibrated subset. The 7D action is decomposed into translation
($p$), rotation ($r$), and gripper ($g$) components, balanced by
homoscedastic uncertainty weighting~\citep{Cipolla18Multi}:
\begin{equation}
  \mathcal{L}_{\mathrm{act}}
  = \sum_{m \in \{p,r,g\}}
    \Bigl(
      e^{-s_m}\,
      \mathrm{SmoothL1}\!\left(\hat{a}_t^{(m)},\, a_t^{(m)}\right)
      + s_m
    \Bigr)
  \label{eq:action_loss}
\end{equation}
with learnable log-variances $s_m = \log \sigma_m^2$ that absorb
relative modality scales during training.

After stage 1, the model serves two roles: it provides a physically
grounded latent action space whose neighborhoods reflect executable
motion rather than visual similarity, and it acts as a fixed tokenizer
that maps real and generated frame pairs to discrete action sequences
for downstream stages.

\subsection{Latent action-guided VLA pretraining}
\label{sec:pretrain}

\begin{figure}[t]
  \centering
  \includegraphics[width=\textwidth]{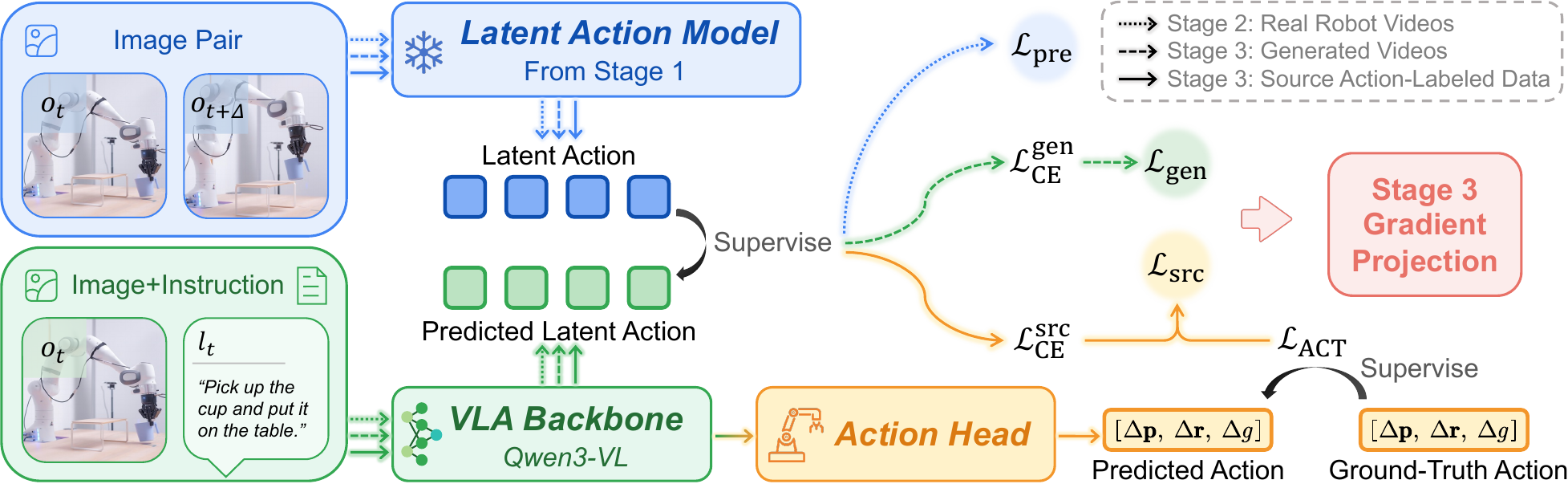}
  \caption{\textbf{Stages~2 and~3 of FreeAct.} The frozen stage-1
  latent action model (top-left) tokenizes a frame pair into a
  discrete sequence, which the VLA backbone is trained to
  autoregressively predict from the current image and instruction
  $\ell_t$. Stage~2 (\textit{dotted}) pretrains the backbone on real
  robot videos with cross-entropy loss
  $\mathcal{L}_{\mathrm{pre}}$. Stage~3 jointly trains two streams:
  source action-labeled data (\textit{solid}), supervising both the
  latent tokens and a continuous action head
  ($\mathcal{L}_{\mathrm{src}}$); and pseudo-labeled generated
  target-environment videos (\textit{dashed}), supervising only the
  latent tokens ($\mathcal{L}_{\mathrm{gen}}$). The two stage-3 streams are
  combined through the gradient projection update of
  \autoref{sec:finetune}.}
  \label{fig:vla_architecture}
\end{figure}

The second stage transfers the motion abstraction into the VLA
backbone (\autoref{fig:vla_architecture}). We augment the backbone's text tokenizer vocabulary with $|\mathcal{C}|$ special action tokens
$\{\texttt{<la}_0\texttt{>}, \dots,
\texttt{<la}_{|\mathcal{C}|-1}\texttt{>}\}$, one per codebook entry,
and cast latent action prediction as next-token prediction within the
backbone's chat template:
\begin{align}
  \texttt{User:} \quad
  & \texttt{<image>}\ \ell_t \notag \\
  \texttt{Assistant:} \quad
  & \texttt{<la}_{z_t^{(1)}}\texttt{>}
    \cdots
    \texttt{<la}_{z_t^{(K)}}\texttt{>}.
    \notag
\end{align}
For each training sample $(o_t, o_{t+\Delta}, \ell_t)$, the frozen stage-1 tokenizer provides the target sequence $z_t = Q_\phi(o_t, o_{t+\Delta})$. The VLA backbone is then trained to
predict it autoregressively:
\begin{equation}
  \mathcal{L}_{\mathrm{pre}}
  = -\sum_{k=1}^{K}
    \log p_\theta\!\left(
      z_t^{(k)} \mid o_t,\, \ell_t,\, z_t^{(<k)}
    \right).
  \label{eq:pretrain_ce}
\end{equation}
The visual encoder is kept frozen throughout; only the
vision--language projector and language model are updated.
This objective scales beyond densely annotated trajectories, since
latent action tokens can be extracted from any unlabeled video. It exposes
the backbone to a physically grounded motion representation before
continuous control is introduced in stage~3.

\subsection{Deployment adaptation via generated videos}
\label{sec:finetune}

In the final stage, we adapt the pretrained VLA to the target
environment by jointly optimizing over action-labeled source data and
generated target-environment videos (right side of \autoref{fig:vla_architecture}).

\paragraph{Continuous action head}
We attach a continuous action head $H_\psi$ that decodes the hidden
states at the latent-action token positions into a continuous robot
action, $\hat{a}_t = H_\psi(h_t)$. $H_\psi$ is built on multihead
attention pooling (MAP)~\citep{Lee19Set}: it pools the latent-token
hidden states into a compact action embedding and projects it to the action space. The head
is trained from scratch in stage~3, while the backbone inherits the
representation from stage~2.

\paragraph{Joint finetuning}
For source samples
$(o_t, o_{t+\Delta}, \ell_t, a_t) \in \mathcal{D}_{\mathrm{src}}$, we
supervise both the latent token sequence and the continuous action:
\begin{equation}
  \mathcal{L}_{\mathrm{src}}
  = \mathcal{L}_{\mathrm{CE}}^{\mathrm{src}}
  + \lambda_a\, \mathcal{L}_{\mathrm{ACT}},
  \qquad
  \mathcal{L}_{\mathrm{ACT}} = \|\hat{a}_t - a_t\|_1.
  \label{eq:src_loss}
\end{equation}
For generated video samples
$(o_t^g, o_{t+\Delta}^g, \ell_t^g) \in \mathcal{D}_{\mathrm{gen}}$,
no real actions are available; we pseudo-label each sample with the
frozen stage-1 tokenizer,
$\tilde{z}_t^g = Q_\phi(o_t^g, o_{t+\Delta}^g)$, and supervise only
the latent token prediction:
\begin{equation}
  \mathcal{L}_{\mathrm{gen}}
  = \lambda_v\, \mathcal{L}_{\mathrm{CE}}^{\mathrm{gen}}.
  \label{eq:gen_loss}
\end{equation}
The two streams thus play asymmetric roles: source data anchors
continuous control fidelity, while generated videos expose the
backbone to how motion manifests under target-domain appearance,
viewpoint, and object variation.

\paragraph{Gradient conflict mitigation}
This asymmetry makes naive joint optimization unstable: the two
streams carry different supervision signals, one continuous and one
token-level, and their gradients frequently conflict. When this
happens, the generated gradient erodes the precise control signal
carried by action labels. To mitigate this, we adopt a gradient
projection update tailored to our two-stream
setting~\citep{Yu20PCGrad}. Let $g_{\mathrm{src}}$ and
$g_{\mathrm{gen}}$ denote the per-parameter gradients from each
stream. Whenever the gradients conflict,
$g_{\mathrm{src}}^\top g_{\mathrm{gen}} < 0$, we project away the
conflicting component
\begin{equation}
  \tilde{g}_{\mathrm{gen}}
  \leftarrow
  g_{\mathrm{gen}}
  - \frac{g_{\mathrm{src}}^\top g_{\mathrm{gen}}}
         {\|g_{\mathrm{src}}\|^2}
    \, g_{\mathrm{src}}
  \label{eq:pcgrad}
\end{equation}
and update with $g_{\mathrm{src}} + \tilde{g}_{\mathrm{gen}}$. This
preserves the continuous control signal from source data while
keeping the target-domain visual gradient alive on parameters where
the two streams agree.


\section{Experiments}

We evaluate FreeAct along the full pipeline from latent action learning to
deployment adaptation. Our experiments are organized around three research
questions:
\begin{itemize}
    \item [\textbf{RQ1}] Does physically grounded supervision yield a
    kinematically structured latent action space, relative to purely
    reconstruction-based baselines? (\autoref{sec:exp_rq1})
    \item [\textbf{RQ2}] Can generated videos alone adapt a pretrained VLA
    policy to unseen target environments without any real
    demonstrations, and does our grounded LAM provide more reliable
    pseudo-supervision than IDM-based and reconstruction-only
    alternatives?
    (\autoref{sec:exp_rq2})
    \item [\textbf{RQ3}] Can the adapted policy further improve through
    self-evolution on its own rollouts, without expert intervention?
    (\autoref{sec:exp_rq3})
\end{itemize}

\subsection{Experimental setup}
\label{sec:exp_setup}

\paragraph{Datasets}
Stages~1 and~2 use DROID~\citep{Khazatsky24DROID} for latent action learning and VLA pretraining; the calibrated subset with camera extrinsics provides $\mathcal{L}_{\mathrm{act}}$ supervision. Stage~3 draws source data from \citet{Liu23LIBERO} and evaluates on a custom robosuite~\citep{Zhu25robosuite} target environment with substantial visual and task shift. Target-domain videos are synthesized with Wan~2.6~\citep{Wan25Wan} (20 rollouts per task).
Full setup details are in \autoref{app:datasets}.

\paragraph{Baselines}
For the latent space evaluation (RQ1), we compare against LAPA~\citep{Ye25LAPA} and UniVLA~\citep{Bu25UniVLA}, two LAMs trained with reconstruction objectives alone. For end-to-end adaptation (RQ2), we compare against UniPi~\citep{Du23UniPi}, which recovers continuous actions from generated videos via an IDM, DreamGen~\citep{Jang25DreamGen}, which finetunes the generator before IDM labeling, and the LAPA and UniVLA tokenizers swapped into our stage-2/3 pipeline.

\paragraph{Implementation details}
Stage~1 uses a frozen DINOv3 ViT-B/16~\citep{Simeoni25DINOv3} as the visual backbone with a transformer encoder--decoder for the VQ-VAE. Stage~2 initializes from Qwen3-VL-4B~\citep{Bai25Qwen3-VL}, keeping the visual encoder frozen and finetuning the projector and language model. Stage~3 inherits the stage-2 checkpoint and jointly trains the continuous action head and backbone on source and generated data. Full hyperparameters and training schedules are in \autoref{app:hyperparams}.

\subsection{RQ1: Physically grounded latent action space}
\label{sec:exp_rq1}

We first evaluate the intrinsic quality of the latent action space
learned in stage~1, asking whether physical action supervision
produces a representation whose geometry reflects executable
kinematics rather than visual texture. 
All metrics below are computed on a held-out DROID test split
disjoint from our training trajectories. Baselines are
LAPA~\citep{Ye25LAPA} and UniVLA~\citep{Bu25UniVLA}, using their official checkpoints; both were trained on broader datasets
that include our test trajectories.

\begin{figure}[t]
  \centering
  \begin{subfigure}[t]{0.3\textwidth}
    \centering
    \includegraphics[width=\textwidth]{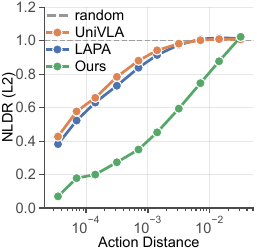}
    \caption{NLDR.}
    \label{fig:nldr}
  \end{subfigure}
  \hfill
  \begin{subfigure}[t]{0.69\textwidth}
    \centering
    \includegraphics[width=\textwidth]{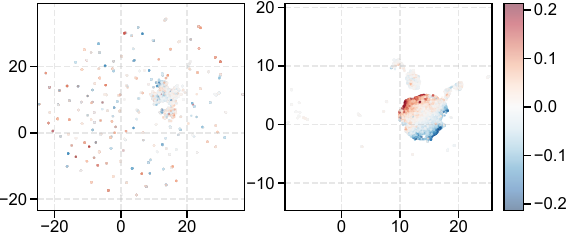}
    \caption{UMAP projection of test-set latents.}
    \label{fig:umap}
  \end{subfigure}
  \caption{\textbf{Latent action space structure on the DROID
  held-out test split.} (a) NLDR vs.\ action-distance threshold
  $\tau$; lower values indicate that physically similar actions are
  colocated in the latent space. (b) UMAP projection of test-set
  latents, colored by end-effector $x$-displacement in meters
  (left: UniVLA; right: ours). LAPA and additional axes are deferred
  to \autoref{sec:additional_results}.}
  \label{fig:latent-structure}
\end{figure}

\paragraph{Topology preservation}
A well-structured latent space should bring physically similar
actions closer than dissimilar ones. We quantify this with a
\emph{normalized latent-distance ratio} (NLDR): for an action-distance
threshold $\tau$, we compare the mean latent distance among sample
pairs whose ground-truth actions are within $\tau$ to the mean
distance over random pairs,
\begin{equation}
\mathrm{NLDR}(\tau)
= \frac{\mathbb{E}_{(i,j) \in \mathcal{S}_\tau} \|z_i - z_j\|_2}
       {\mathbb{E}_{\mathrm{random}} \|z_u - z_v\|_2},
\qquad
\mathcal{S}_\tau = \{(i, j) : d_a(a_i, a_j) < \tau\}.
\end{equation}
The action distance $d_a$ sums the mean SmoothL1 distance over the
three semantic components of the 7D action: translation ($p$),
rotation ($r$), and gripper ($g$), giving each component equal
weight regardless of its dimensionality. A value near $1$ indicates
no correlation between latent and physical distance; values near $0$
indicate that physically similar actions form tight latent clusters.
As shown in \autoref{fig:nldr}, our model reaches NLDR below $0.1$
at the smallest $\tau$ and stays well below the baselines across the
full range; LAPA and UniVLA improve only marginally over random as $\tau$ shrinks, with NLDR remaining around 0.4 even at the tightest threshold.

\begin{wrapfigure}{r}{0.36\textwidth}
  \centering
  \vspace{-18pt}
  \includegraphics[width=0.36\textwidth]{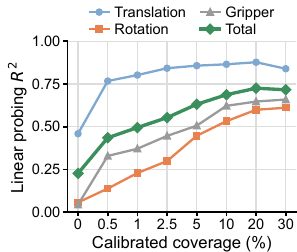}
  \vspace{-6pt}
  \caption{\textbf{Linear probing $R^2$ vs.\ action-supervision
  coverage.} Stage~1 is retrained with $\mathcal{L}_{\mathrm{act}}$
  restricted to nested subsets of calibrated trajectories.}
  \label{fig:coverage_sweep}
  \vspace{-20pt}
\end{wrapfigure}

\paragraph{Manifold structure}
\autoref{fig:umap} projects test-set latents into 2D~\citep{Yuan21VASurvey} via
UMAP~\citep{McInnes18UMAP}, colored by ground-truth end-effector
$x$-displacement. The UniVLA latent space disperses into a noisy
central cluster in which colors are intermixed with no discernible
gradient: the continuous action space is fractured into unrelated
fragments, leaving no geometric reference along which a downstream
policy can interpolate. Our model instead forms a dense manifold
whose color varies smoothly from one side to the other, mapping the
topology of physical motion onto the latent space. Adjacent latent
vectors thus correspond to neighboring physical actions, the
prerequisite for fine-grained control and smooth skill generalization in downstream stages.

\paragraph{Linear decodability}
A continuous manifold structure is only useful if downstream modules
can read off action information from it. We probe this with ridge
regression from frozen latents to the 7D ground-truth action,
reporting the coefficient of determination $R^2$ (higher is better)
on train and test splits. \autoref{tab:linear_probe} shows that LAPA
and UniVLA achieve $R^2 \!\approx\! 0.1$ overall, with rotation and
gripper components effectively undecodable ($R^2 \!<\! 0.03$). Our
model reaches $R^2 = 0.76$ overall, with translation, rotation, and
gripper all linearly decodable in their own right. The latent space
not only clusters by kinematics, it explicitly encodes them.

\paragraph{Supervision coverage}
Camera extrinsics are available for only $\sim$40\% of DROID,
so we vary action-supervision coverage to assess how much calibrated
data grounding requires. We retrain stage~1 with the action loss
restricted to nested subsets of calibrated trajectories, keeping
all other data and settings fixed
(\autoref{fig:coverage_sweep}). Within this sweep, reducing coverage
from 30\% to 0\% ($\lambda_{\mathrm{act}}=0$) lowers $R^2$ from
0.72 to 0.23, with rotation and gripper falling from 0.61/0.66 to
0.06/0.04. Performance levels off around 20\% coverage; 1\% already
attains 69\% of the plateau value, indicating that limited
calibration can substantially improve action decodability.

\begin{table}[t]
\centering
\caption{\textbf{Linear probing $R^2$ on the DROID held-out test set.}
Reconstruction-only baselines fail on rotation and gripper dimensions;
our model achieves high decodability across all components.}
\label{tab:linear_probe}
\setlength{\tabcolsep}{12pt}
\begin{tabular}{lcccccc}
\toprule
\multirow{2}{*}{\textbf{Method}}
& \multicolumn{2}{c}{\textbf{LAPA}}
& \multicolumn{2}{c}{\textbf{UniVLA}}
& \multicolumn{2}{c}{\textbf{Ours}} \\
\cmidrule(lr){2-3} \cmidrule(lr){4-5} \cmidrule(l){6-7}
& Train & Test & Train & Test & Train & Test \\
\midrule
Total $R^2$         & 0.104 & 0.107 & 0.096 & 0.090 & 0.794 & \textbf{0.761} \\
Translation $R^2$   & 0.223 & 0.226 & 0.201 & 0.189 & 0.899 & \textbf{0.878} \\
Rotation $R^2$      & 0.013 & 0.014 & 0.016 & 0.013 & 0.711 & \textbf{0.660} \\
Gripper $R^2$       & 0.022 & 0.029 & 0.019 & 0.021 & 0.727 & \textbf{0.712} \\
\midrule
Norm.\ MSE          & 0.896 & 0.990 & 0.904 & 1.008 & 0.206 & \textbf{0.264} \\
\bottomrule
\end{tabular}
\end{table}

Across these diagnostics, the picture is consistent: physically
grounded supervision through the auxiliary Action Decoder is
sufficient to convert a reconstruction-only codebook into a
continuous, linearly decodable kinematic manifold. This structural
property is what allows downstream stages to convert generated
videos into reliable adaptation supervision (\autoref{sec:exp_rq2}).

\subsection{RQ2: Demonstration-free adaptation from generated videos}
\label{sec:exp_rq2}

\begin{figure}[t]
  \centering

  \begin{subfigure}[t]{0.24\textwidth}
    \centering
    \includegraphics[width=\linewidth]{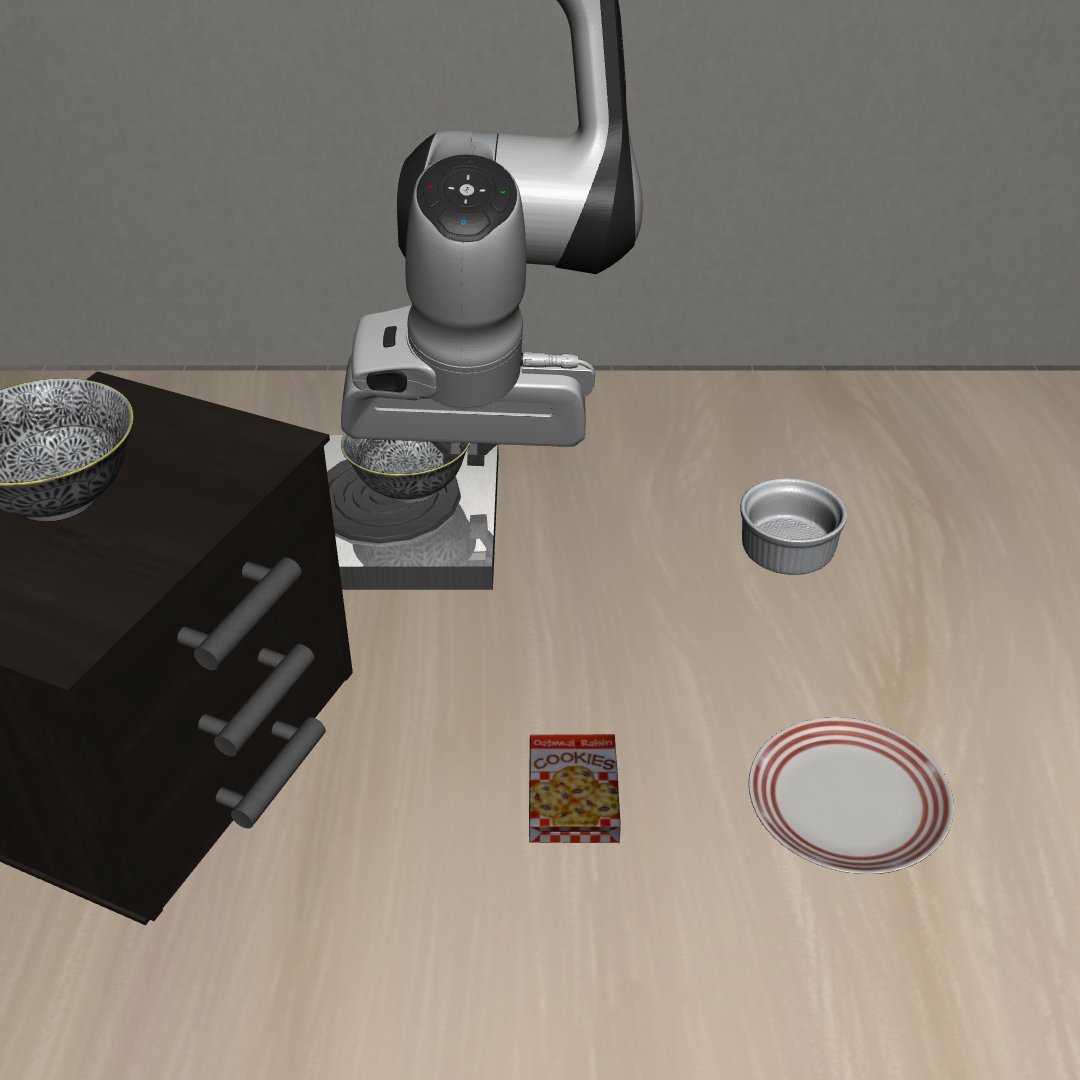}
    \caption{Source Env}
    \label{fig:catch_original}
  \end{subfigure}
  \hfill
  \begin{subfigure}[t]{0.24\textwidth}
    \centering
    \includegraphics[width=\linewidth]{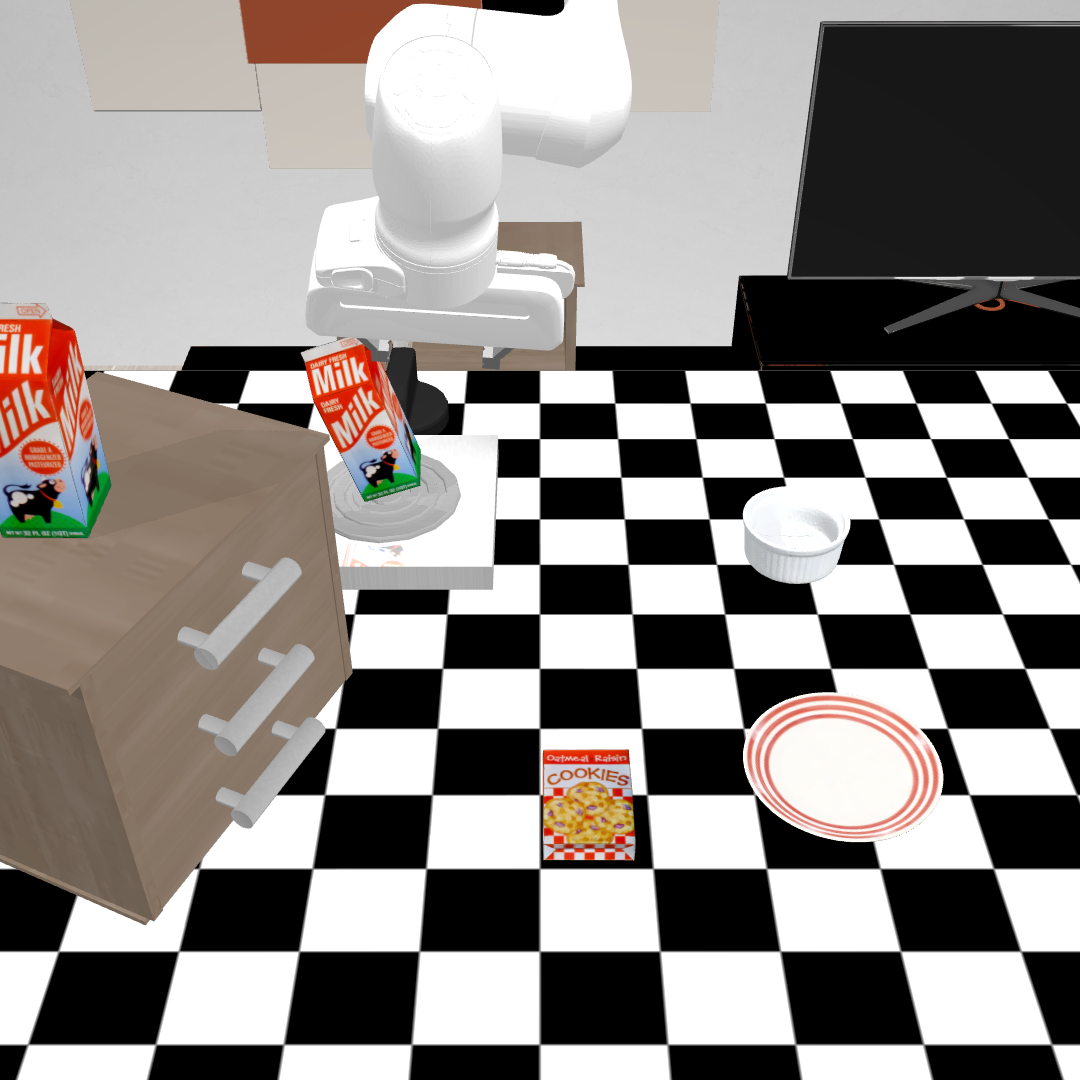}
    \caption{Target Env}
    \label{fig:catch_fail}
  \end{subfigure}
  \hfill
  \begin{subfigure}[t]{0.24\textwidth}
    \centering
    \includegraphics[width=\linewidth]{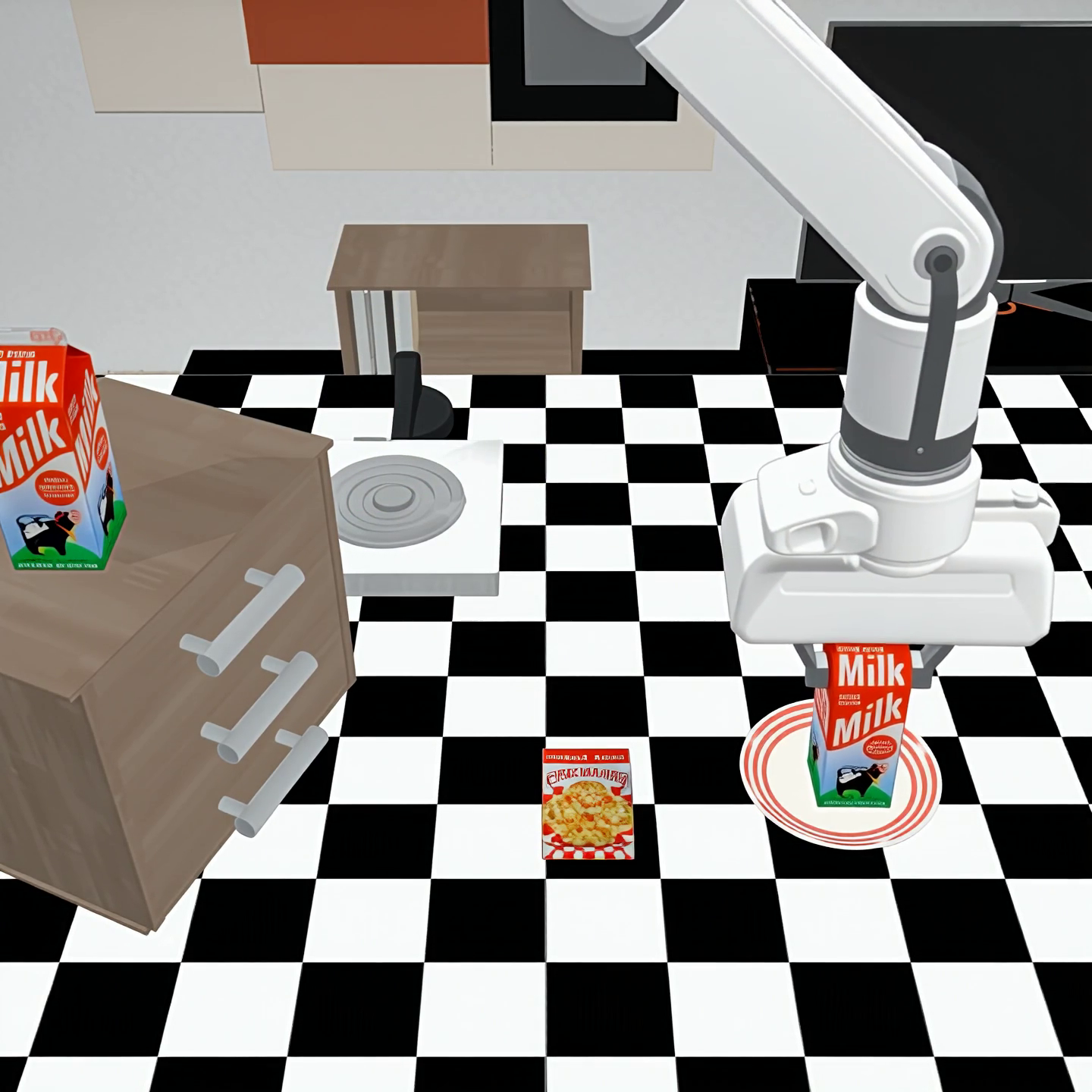}
    \caption{Generated Video}
    \label{fig:last_frame_gen}
  \end{subfigure}
  \hfill
  \begin{subfigure}[t]{0.24\textwidth}
    \centering
    \includegraphics[width=\linewidth]{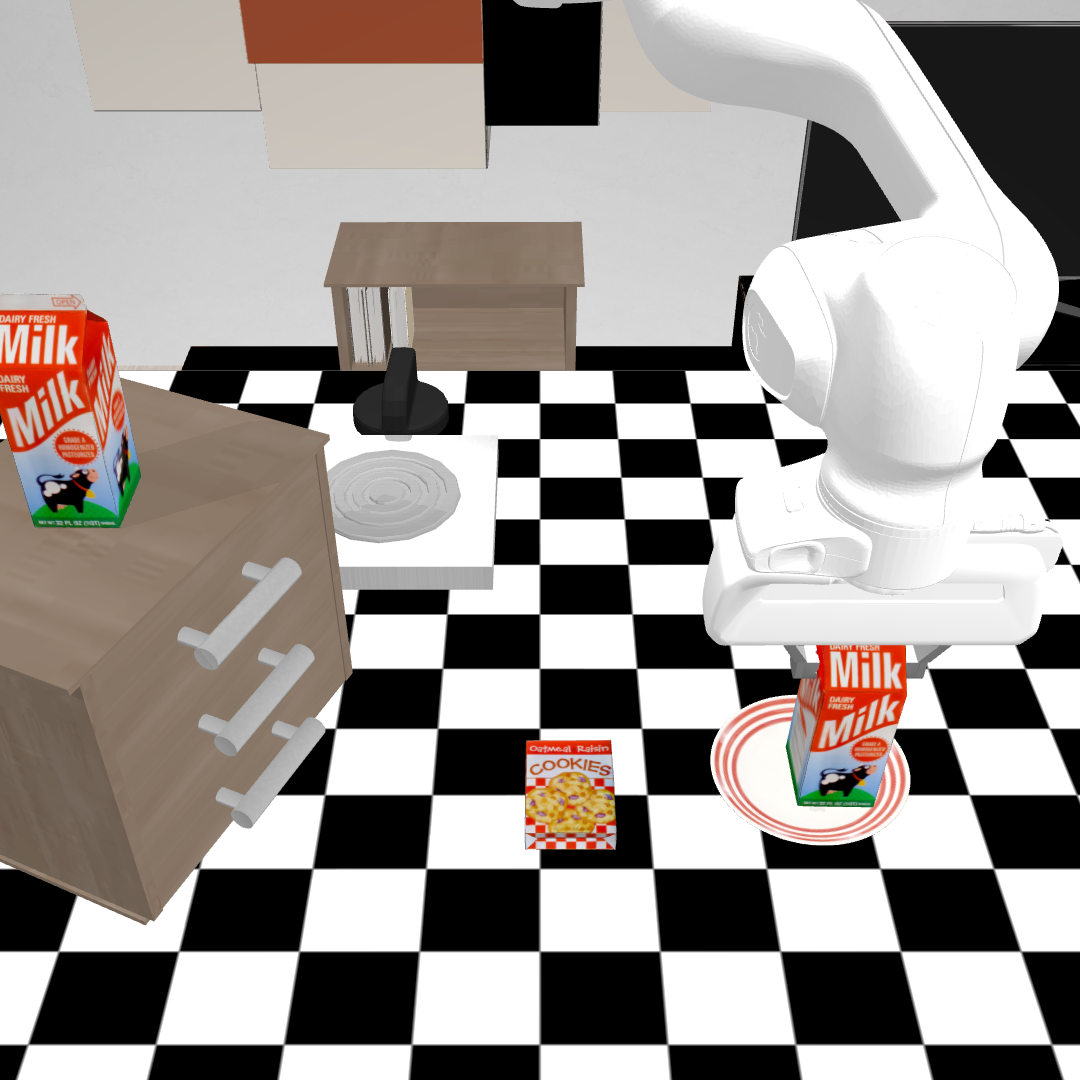}
    \caption{Rollout}
    \label{fig:last_frame}
  \end{subfigure}

  \caption{\textbf{An example of the source--target shift, generated supervision, and resulting rollout (Task E).} (a) Source environment. (b) Target environment, exemplifying the visual and task changes considered in this paper; without adaptation, the source policy topples the milk carton with its pinch trajectory. (c) Frame from a Wan~2.6 generated rollout: the arm shows kinematic hallucinations, yet the gripper and end-effector remain faithful. (d) Successful FreeAct rollout after adaptation.}
  \label{fig:env_comparison}
\end{figure}

We now turn to the central question of this paper: can a pretrained
VLA policy be adapted to a new environment using only generated
videos, with zero expert demonstrations? We construct a target
environment with changes in lighting, desk textures, robot link colors, initial
joint configurations, task instructions, and most importantly the
target manipulation skill itself (\autoref{fig:env_comparison}a, b).
Source tasks reduce to a \emph{precision pinch} on a thin protrusion,
exemplified by the rim of a bowl, where the policy only needs to
align one finger with a narrow edge before closing. Target tasks
instead require an \emph{enclosing grasp} on a milk carton, an object
never seen in the source distribution: the gripper must approach
with both fingers straddling the full width of the body, at a
different height, direction, and closing pattern. The two regimes
thus exercise different aspects of low-level control, not merely
different appearances.

\begin{figure}[t]
  \centering
  \includegraphics[width=\textwidth]{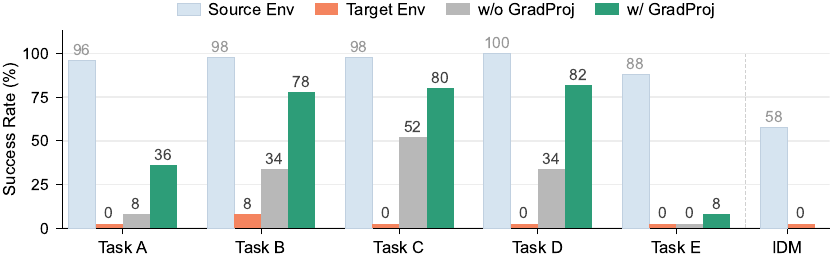}
  \caption{\textbf{Demonstration-free adaptation results on five
  target tasks.} \emph{Source Env}: pretrained VLA on the source
  environment (sanity check). \emph{Target Env}: same policy deployed
  on the target environment without adaptation. \emph{w/ GradProj}
  (FreeAct): adaptation using only Wan~2.6 generated videos
  pseudo-labeled by our latent action model, with the gradient
  projection update of \autoref{sec:finetune}. \emph{w/o GradProj}:
  the same setup with the projection removed. \emph{IDM}:
  UniPi-style baseline, evaluated on real source videos (left, $58\%$)
  and on generated target videos (right, $0\%$). Each bar averages 50
  closed-loop rollouts.}
  \label{fig:zero-shot_result}
\end{figure}

\autoref{fig:zero-shot_result} reports closed-loop success across
five target tasks, averaged over 50 trials each. The pretrained VLA,
despite near-saturated source performance ($96\%$ on average),
collapses to $1.6\%$ when deployed directly on the target
environment. Lacking any prior exposure to carton-shaped objects,
the policy keeps executing its source-domain pinch trajectory,
swinging one finger of the parallel gripper into the carton and
toppling it (\autoref{fig:catch_fail}). FreeAct, supervised
only by Wan~2.6 generated videos
(\autoref{fig:last_frame_gen}), recovers an average success rate
of $57\%$ (\autoref{fig:last_frame}), with four of five
tasks exceeding $35\%$ and three exceeding $75\%$.

A natural alternative is to recover actions directly from generated
videos via IDM, as in UniPi. We evaluate it on two inputs separately.
On real teleoperated source-environment videos the IDM is faithful,
replaying its predictions with $58\%$ success and confirming our
reimplementation is sound. On Wan~2.6 generated videos the same IDM
collapses to $0\%$: kinematic inconsistencies push the inputs
outside its training distribution, and the recovered action sequences
become physically incoherent.
DreamGen's official pipeline, which
uses a robot-finetuned video generator followed by IDM labeling,
also yields $0\%$ success.
While the generated arm is
hallucinated, the gripper and end-effector motion remain faithful
to the task; our latent action model is designed to extract
exactly this signal (\autoref{sec:lam}).
For comparison, we use the
official reconstruction-only LAPA or UniVLA tokenizer in the same
stage-2/3 pipeline. Source success remains high ($87\%$ and $89\%$),
but target success reaches only $12\%$ and $4\%$
(\autoref{tab:tokenizer_swap}), consistent with their lower action
decodability in RQ1.

Removing the gradient projection step (w/o GradProj in
\autoref{fig:zero-shot_result}) drops the average success rate from
$57\%$ to $26\%$, with three of five tasks losing more than
half of their success and Task E collapsing to $0\%$. This confirms
the prediction of \autoref{sec:finetune}: a naive sum of the two
losses lets the generated-video gradient cancel parts of the
source-stream update, eroding the precise control signal carried
by action labels. The projected update strips the conflicting
component from the generated gradient, so source data continues to
deliver clean control supervision while the generated stream still
propagates target-domain visual information through the parameters
where the two streams agree.

Task E remains the hardest case across all settings, including for
FreeAct ($8\%$). The cause traces to the target scene geometry: the
milk carton sits almost directly beneath the gripper and slightly
farther from the camera, weakening the perspective cues that
disambiguate depth. Wan~2.6 interprets this as a vertical descent
and generates videos in which the gripper drops straight down,
ignoring the small forward displacement required for accurate
grasping. The downstream policy is therefore upper-bounded by the
fidelity of the generated supervision: when the generator misreads
the scene geometry, no amount of action grounding can recover the
missing kinematic content. This points toward generation-side
improvements such as multi-view conditioning or depth priors~\citep{Xu26DepthFoundation} as a
complementary axis to action grounding.

\subsection{RQ3: Self-evolution from policy rollouts}
\label{sec:exp_rq3}

The adapted policy from RQ2 recovers basic task competence from
generated videos alone. We now ask whether it can further improve
through its own experience. We deploy the stage-3 policy in the
target environment, score each rollout with GPT-5.5~\citep{OpenAI26GPT55} on task completion and execution
quality (prompt in \autoref{app:mllm_prompt}), and finetune on up
to five of the highest-scoring rollouts per task, drawn from the 50
evaluation rollouts per task in \autoref{fig:zero-shot_result}.
Ground-truth success labels are not used for selection; a post hoc
check confirms that all selected rollouts are successful.

\begin{wrapfigure}{r}{0.3\textwidth}
  \centering
  \vspace{-11pt}
  \includegraphics[width=0.3\textwidth]{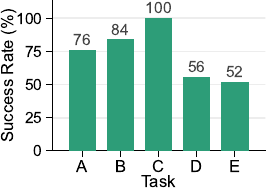}
  \caption{\textbf{Self-evolution from rollouts.} Success rates
  after one round of self-evolution. Each
  bar averages 50 closed-loop rollouts.}
  \label{fig:rq3_result}
  \vspace{-20pt}
\end{wrapfigure}

\autoref{fig:rq3_result} reports success rates after one round of
self-evolution. The average rises from $57\%$ to $74\%$. Task~E, previously capped at
$8\%$ due to depth ambiguity in the generated videos, jumps to
$52\%$, confirming that even minimal real interaction can correct
biases that generated supervision alone cannot resolve. Task~D falls from $82\%$ to $56\%$ despite
successful selected rollouts. Pooling data across tasks may introduce
interference during finetuning, a possible explanation for this decline.

No target-domain expert demonstration is introduced at any point:
stage~3 adapts from generated videos, and self-evolution draws on
the policy's own rollouts as judged by an MLLM. This establishes a
complete adaptation loop for deployment to novel environments,
in which generated videos bootstrap competence and autonomous
refinement lifts performance to practical levels.


\section{Conclusion}

We presented FreeAct, a demonstration-free adaptation framework that
anchors a discrete latent action space to camera-frame end-effector
kinematics, enabling reliable pseudo-labeling of generated videos
despite their kinematic artifacts. Generated videos alone lift a
pretrained VLA from $2\%$ to $57\%$ under aggressive domain shift;
the adapted policy then self-evolves on its own rollouts to $74\%$.
These results establish a complete adaptation loop that requires no
target-domain expert demonstrations.

\paragraph{Limitations and future work}
Our evaluation is confined to simulation, a single embodiment, and
single-arm tabletop manipulation; real robots, cross-embodiment
transfer, and bimanual or dexterous settings with richer contact
dynamics are natural next steps. Adaptation is bounded by generator
fidelity (Task~E), which multi-view conditioning or depth priors
could improve. Pooled self-evolution can also reduce performance on
individual tasks (Task~D); per-task balancing is a possible remedy.

\begin{ack}
This work was supported by Fundamental and Interdisciplinary Disciplines Breakthrough Plan of the Ministry of Education of China (No. JYB2025XDXM101), the National Natural Science Foundation of China (project No. 62220106003, 62495061, 623B2057).
\end{ack} 


{
\small
\bibliographystyle{plainnat}
\bibliography{references}
}


\appendix

\section{Hyperparameters and Training Details}
\label{app:hyperparams}

\autoref{tab:hp_stage1}, \autoref{tab:hp_stage2}, and
\autoref{tab:hp_stage3} list the full hyperparameter configurations
for the three training stages. All experiments are run on a single
node with 8 NVIDIA RTX 4090 GPUs (24\,GB each). End-to-end wall-clock
training time is approximately 19 hours for stage 1, 120 hours for
stage 2, and 7 hours for stage 3.

\begin{table}[H]
\centering
\caption{Stage 1 (LAM Training) hyperparameters.}
\label{tab:hp_stage1}
\begin{tabular}{ll}
\toprule
\textbf{Component} & \textbf{Setting} \\
\midrule
Visual backbone & DINOv3 ViT-B/16 (frozen) \\
Encoder / decoder & 12 layers, dim 768, 12 heads \\
VQ codebook size $C$ & 64 \\
Latent dimension & 128 \\
Tokens per frame pair $K$ & 4 \\
Action Decoder & 3-layer MLP, hidden $[256, 128]$, GELU \\
\midrule
Training data & DROID \\
Global batch size & 512 (64 per GPU) \\
Optimizer & AdamW (lr $10^{-4}$, wd $10^{-2}$) \\
Gradient clipping & 0.1 \\
Epochs & 1 \\
Action loss weight $\lambda_{\mathrm{act}}$ & 0.01 \\
\bottomrule
\end{tabular}
\end{table}

\begin{table}[H]
\centering
\caption{Stage 2 (VLA Pretraining) hyperparameters.}
\label{tab:hp_stage2}
\begin{tabular}{ll}
\toprule
\textbf{Component} & \textbf{Setting} \\
\midrule
Backbone & Qwen3-VL 4B-Instruct \\
Visual encoder & frozen \\
Projector & full precision, fully trained \\
Language model & LoRA (rank 64, $\alpha = 128$) \\
\midrule
Global batch size & 256 \\
Optimizer & AdamW (lr $5{\times}10^{-5}$, wd $10^{-2}$) \\
Schedule & constant with 3000-step warmup \\
Parallelism & DeepSpeed ZeRO-2 \\
Epochs & 2 \\
Latent tokenization & on-the-fly (lazy encoding) \\
\bottomrule
\end{tabular}
\end{table}

\begin{table}[H]
\centering
\caption{Stage 3 (Deployment Adaptation) hyperparameters.}
\label{tab:hp_stage3}
\begin{tabular}{ll}
\toprule
\textbf{Component} & \textbf{Setting} \\
\midrule
Initialization & Stage 2 checkpoint \\
Action head hidden dim & 512 \\
Backbone learning rate & $10^{-4}$ (cosine decay to 10\% of peak) \\
Action head learning rate & $3.5{\times}10^{-4}$ (wd $10^{-3}$) \\
Global batch size & 256 \\
Epochs & 75 \\
Warmup steps & 200 \\
Generated data weight $\lambda_v$ & 0.1 \\
Action window $W$ & 16 \\
Gradient projection & enabled at every step \\
\bottomrule
\end{tabular}
\end{table}

\section{Data and Task Details}
\label{app:datasets}

\paragraph{DROID (Stages~1 and~2)}
The DROID dataset provides approximately 362 hours of teleoperated manipulation across 564 scenes. The $\sim$40\% of trajectories with precise camera extrinsics form the calibrated subset on which the action loss $\mathcal{L}_{\mathrm{act}}$ is applied; the remaining trajectories contribute only to the reconstruction and VQ objectives.

\paragraph{LIBERO (Stage~3 source)}
Source data $\mathcal{D}_{\mathrm{src}}$ is drawn from LIBERO~\citep{Liu23LIBERO}. We use its standard task suite as the source-domain action-labeled data for joint finetuning.

\paragraph{Target environment (Stage~3)}
The target environment is a custom robosuite~\citep{Zhu25robosuite} scene that differs from the source in lighting, desk textures, object identities and layout, and task instruction. Target-domain videos $\mathcal{D}_{\mathrm{gen}}$ are synthesized with Wan~2.6~\citep{Wan25Wan}, conditioned on the initial frame and the language instruction, yielding 20 rollouts per task.

\paragraph{Target task suite}
The five target tasks instantiate a single unseen manipulation skill
(enclosing power grasp followed by short-range relocation) under
systematically varied spatial contexts. This controlled design is deliberate:
by fixing the grasp type and success criterion, we isolate how well
generated-video supervision generalizes across different approach geometries
without confounding from semantic task variation. The five variants differ in
support surface, surrounding distractors, and spatial relation between the
target object and nearby landmarks, producing meaningfully different
trajectory requirements (collision avoidance, depth estimation, pre-grasp
alignment). Table~\ref{tab:target_tasks} lists the instruction templates.

\begin{table}[H]
\centering
\caption{Target task suite. Each task instantiates the same unseen-object
relocation skill under a different spatial configuration, varying the approach
geometry, support relation, and surrounding distractors while keeping the
target object and success criterion controlled.
\texttt{[obj]}, \texttt{[ref]}, and \texttt{[goal]} denote the target object,
reference landmark(s), and goal location, respectively.}
\label{tab:target_tasks}
\begin{tabular}{@{}clp{7.8cm}@{}}
\toprule
Task & Spatial context & Instruction template \\
\midrule
A & Between distractors & Pick up \texttt{[obj]} between \texttt{[ref\textsubscript{1}]} and \texttt{[ref\textsubscript{2}]} and place it on \texttt{[goal]}. \\
B & Open table surface & Pick up \texttt{[obj]} from \texttt{[ref]} and place it on \texttt{[goal]}. \\
C & On supporting object & Pick up \texttt{[obj]} on \texttt{[ref]} and place it on \texttt{[goal]}. \\
D & Adjacent to distractor & Pick up \texttt{[obj]} next to \texttt{[ref]} and place it on \texttt{[goal]}. \\
E & Depth-ambiguous region & Pick up \texttt{[obj]} on \texttt{[ref]} and place it on \texttt{[goal]}. \\
\bottomrule
\end{tabular}
\end{table}

\section{MLLM Evaluation Prompt}
\label{app:mllm_prompt}
In RQ3 (\autoref{sec:exp_rq3}), we use GPT-5.5 as a rollout evaluator. The model receives the full rollout video along
with the task instruction and returns a composite score (five
dimensions, each 0--4, totaling 20 points) assessing task completion
and execution quality. The prompt template is shown below.

\begin{lstlisting}[style=prompt]
You will watch a rollout video of a robot completing the <task objective> and score it based on the quality of the task execution process.
Please inspect the video frame by frame. Do not score based only on key frames, sampled frames, or an overall impression. Before scoring, you must clearly observe the continuous process from the robot approaching the target object, making contact, performing the operation, and reaching the end of the task. Pay close attention to the approach path, pauses, jitter, incorrect contact, corrective actions, environmental collisions, operation effect, and final state.
Please score the performance across the following five dimensions. Each dimension is scored from 0 to 4, for a total of 20 points. Each score must be an integer and include a brief reason.

(1) Approach and Initial Operation Smoothness
Evaluate whether the robot's process from approaching the target object to forming effective contact is natural, continuous, and accurate.
4 points: The robot approaches accurately in one attempt and forms effective contact. The path is natural and continuous, with no obvious jitter, pauses, probing, incorrect contact, or secondary adjustment.
3 points: Overall fairly smooth, with only slight deceleration, small pose correction, or very brief alignment. No obvious pause, hesitation, jitter, poking, pushing, retreating, or post-contact readjustment.
2 points: Average process, with visible pauses, hesitation, jitter, path deviation, obvious pose correction, brief incorrect stopping, slight poking, pushing, or post-contact adjustment. Effective contact is eventually formed.
1 point: Effective contact is formed only after multiple pokes, pushes, probing actions, retreats, or repeated adjustments.
0 points: The robot is unable to form effective contact for a long time, repeatedly collides with or pushes away the target object, or makes incorrect contact; the initial stage is chaotic.

(2) Contact Position Accuracy and Operation Quality
Evaluate whether the robot's contact position, pose, contact method, force direction, and stability are reasonable.
4 points: The contact position is ideal, the pose and force application are reasonable, contact is stable and sufficient, and the target object shows no obvious sliding, rotation, tilting, or instability.
3 points: The contact position is generally accurate, with only slight offset or slightly insufficient contact. The target object remains basically stable, with no obvious sliding, rotation, tilting, slipping, or instability risk.
2 points: The contact position is not ideal, with obvious offset, insufficient contact area, poor force direction, or uneven force, causing visible tilting, sliding, rotation, slight slipping, or instability of the target object. The task can still continue.
1 point: The contact position is poor, contacting only an edge, a small area, or the wrong area. The target object is clearly unstable, nearly drops, gets stuck, or fails.
0 points: The contact position is extremely poor, causing clear operation failure, severe instability, or requiring recovery from a failed state.

(3) Execution Stability and Trajectory Smoothness
Evaluate whether the main execution phase is stable, smooth, and continuous. Do not separately evaluate environmental collisions here.
4 points: The execution process is continuous and natural, with steady speed and no obvious jitter, sudden stops, retreats, detours, or unnecessary adjustments. The target object and end effector remain stable.
3 points: Overall stable, with only slight speed changes, small trajectory adjustments, very brief pauses, or slight shaking. No obvious pose changes, instability, or unnecessary corrections.
2 points: The process is not very smooth, with obvious pauses, jitter, uneven speed, path corrections, pose adjustments, repeated micro-adjustments, brief retreats, or unnecessary detours. The target object or end effector visibly shakes, tilts, or becomes unstable, but the task can still continue.
1 point: Multiple jitters, sudden stops, retreats, or repeated adjustments occur. The trajectory is clearly unnatural, and the target object or end effector is obviously unstable.
0 points: The execution process is chaotic, with long stagnation, severe jitter, or multiple ineffective movements. The task state is severely unstable or close to losing control.

(4) Environmental Interaction and Collision Control
Evaluate whether the robot, end effector, or target object makes unnecessary contact with the environment and causes disturbances.
4 points: No obvious collision or unnecessary contact throughout the process; no non-target objects or environmental structures are disturbed.
3 points: Only slight, brief scraping or near-contact occurs, without visible movement, shaking, squeezing, dragging, mistaken contact, or task interference.
2 points: Obvious but not severe collision, scraping, pressing, dragging, or mistaken contact occurs, causing small movement, shaking, pose change, brief obstruction, or slight interference, but does not lead to failure.
1 point: The robot collides with environmental objects multiple times, or clearly pushes, squeezes, drags, or mistakenly contacts objects, causing noticeable disturbance or affecting the task.
0 points: Severe collision or major environmental disturbance occurs, causing objects to be knocked away, dragged, toppled, mistakenly triggered, or seriously interfering with the task.

(5) Task Success and Final State Accuracy
Evaluate whether the task is completed and whether the final result reaches the target position or target state.
4 points: The task succeeds. The final result is accurately in place and stable, with no obvious deviation, over-operation, or risk of failure.
3 points: The task succeeds. The final result is basically in place, with only slight position or pose deviation. The target object is stable, with no obvious edge offset, tilting, slipping, or subsequent instability risk.
2 points: The task is basically successful, but the final position, pose, or state is clearly not ideal, such as obvious deviation from the target center, edge offset, tilting, incomplete placement, or risk of slipping, toppling, deviation, rebound, or later instability.
1 point: The task barely succeeds or is close to succeeding. The final state is unstable, clearly deviates from the target, or only part of the objective is completed.
0 points: The task fails. The target position or state is not reached, or dropping, non-triggering, severe deviation, jamming, instability, or another issue prevents task completion.

Please output in the following format:
(1) Approach and Initial Operation Smoothness: X/4
Reason: ...
(2) Contact Position Accuracy and Operation Quality: X/4
Reason: ...
(3) Execution Stability and Trajectory Smoothness: X/4
Reason: ...
(4) Environmental Interaction and Collision Control: X/4
Reason: ...
(5) Task Success and Final State Accuracy: X/4
Reason: ...
Total Score: X/20
\end{lstlisting}

\section{Additional Results}\label{sec:additional_results}

\autoref{tab:tokenizer_swap} gives the per-task results of the
tokenizer-swap experiment of \autoref{sec:exp_rq2}.

\begin{table}[H]
\centering
\caption{\textbf{Tokenizer swap.} Success rate (\%) when the
official LAPA and UniVLA latent action models replace ours in an
otherwise identical stage-2/3 pipeline on the same training data.
Each entry averages 50 closed-loop rollouts.}
\label{tab:tokenizer_swap}
\setlength{\tabcolsep}{6pt}
\begin{tabular}{llcccccc}
\toprule
& Tokenizer & A & B & C & D & E & Avg \\
\midrule
\multirow{3}{*}{Source Env}
& UniVLA & 74 & 96 & 100 & 94 & 80 & 88.8 \\
& LAPA   & 78 & 84 & 92  & 92 & 90 & 87.2 \\
& Ours   & 96 & 98 & 98  & 100 & 88 & \textbf{96.0} \\
\midrule
\multirow{3}{*}{Target Env (adapted)}
& UniVLA & 14 & 4  & 0  & 0  & 2 & 4.0 \\
& LAPA   & 10 & 32 & 18 & 0  & 0 & 12.0 \\
& Ours   & 36 & 78 & 80 & 82 & 8 & \textbf{56.8} \\
\bottomrule
\end{tabular}
\end{table}

\begin{figure}[H]
  \centering
  \includegraphics[width=\textwidth]{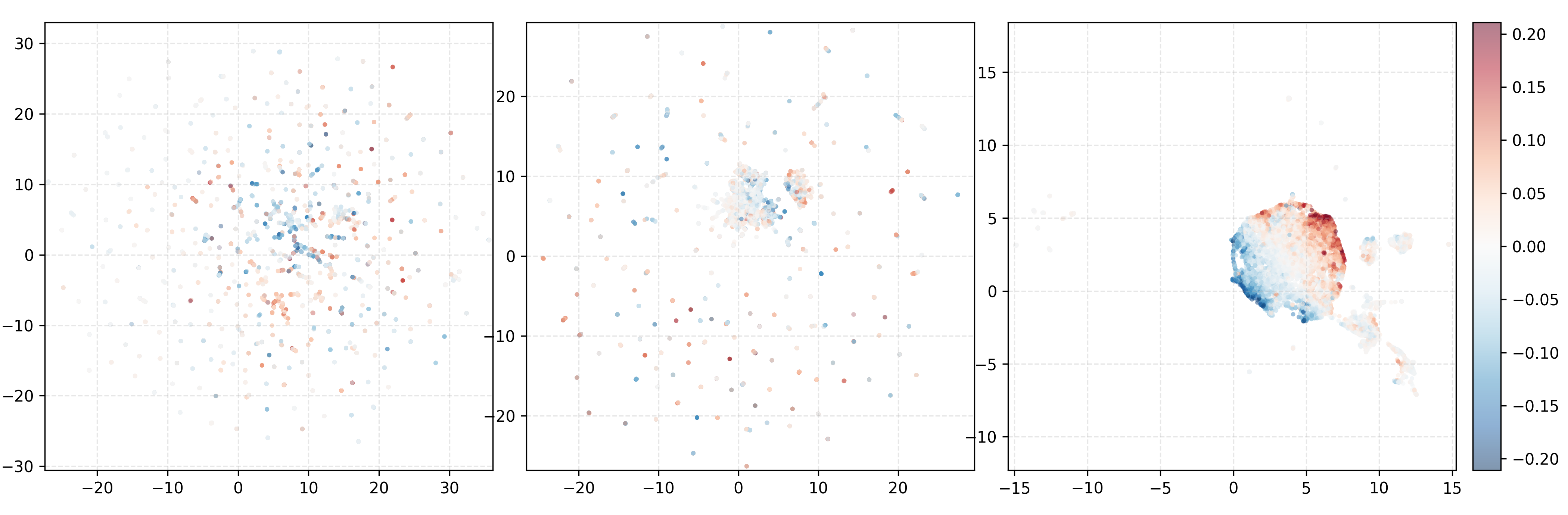}
  \caption{UMAP visualization colored by X-axis displacement. \textbf{Left}: LAPA; \textbf{Middle}: UniVLA; \textbf{Right}: Ours.}
  \label{fig:umap_x}
\end{figure}

\begin{figure}[H]
  \centering
  \includegraphics[width=\textwidth]{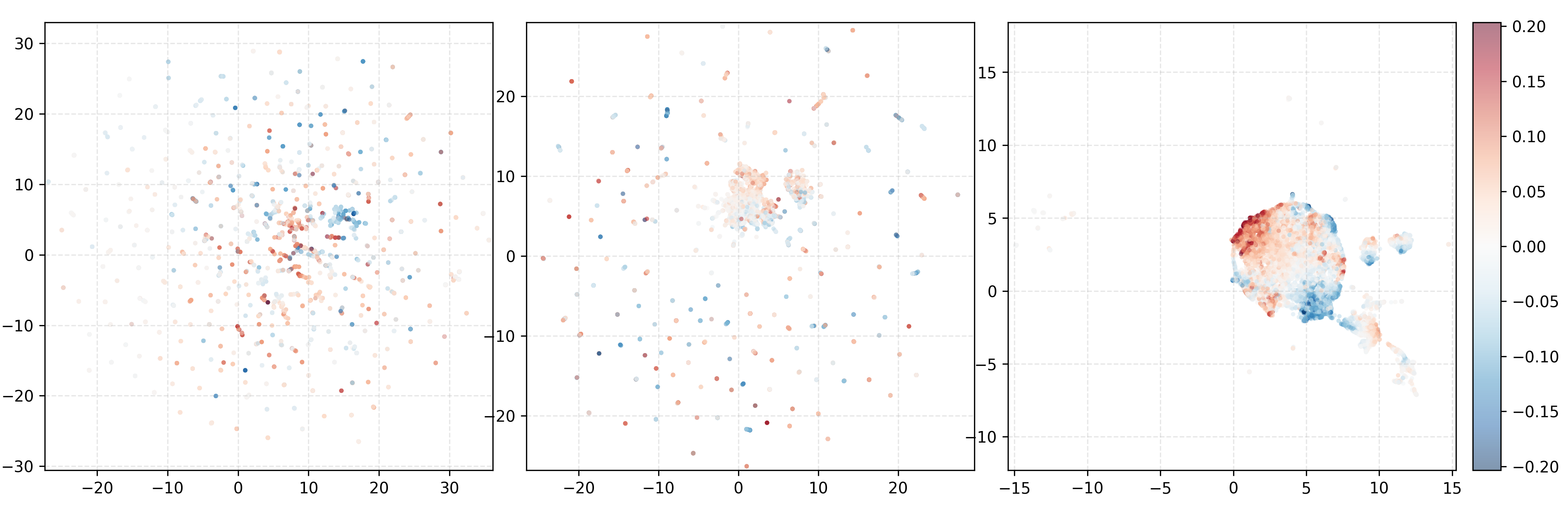}
  \caption{UMAP visualization colored by Y-axis displacement. \textbf{Left}: LAPA; \textbf{Middle}: UniVLA; \textbf{Right}: Ours.}
  \label{fig:umap_y}
\end{figure}

\begin{figure}[H]
  \centering
  \includegraphics[width=\textwidth]{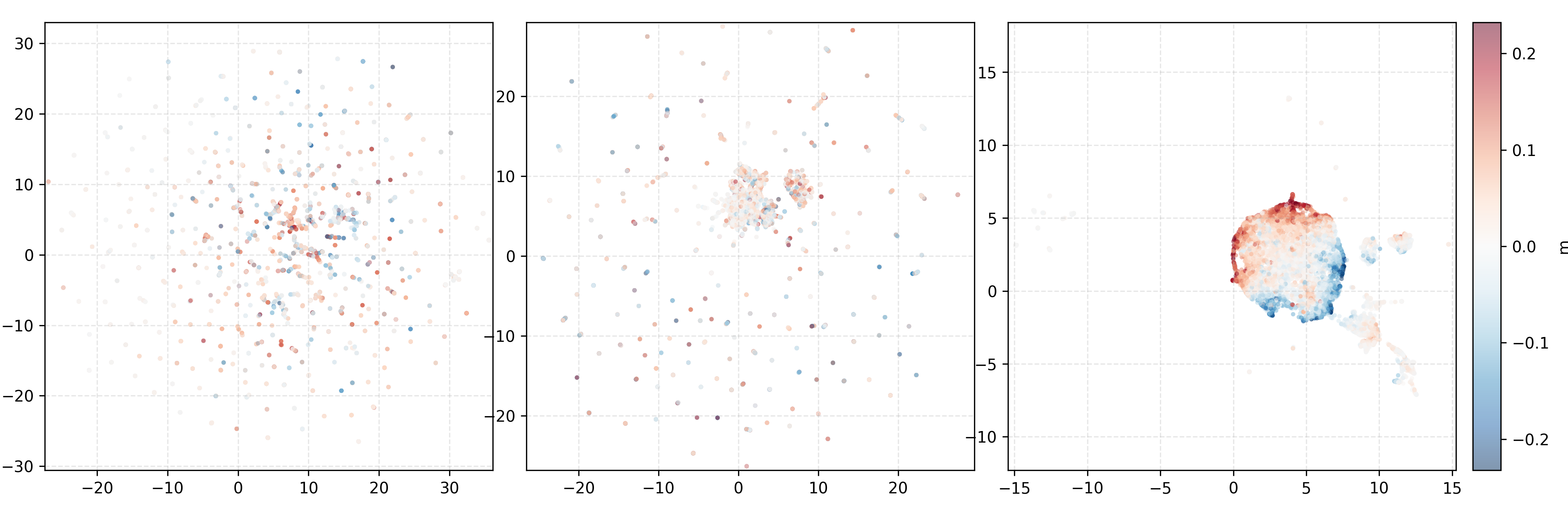}
  \caption{UMAP visualization colored by Z-axis displacement. \textbf{Left}: LAPA; \textbf{Middle}: UniVLA; \textbf{Right}: Ours.}
  \label{fig:umap_z}
\end{figure}

\begin{figure}[H]
  \centering
  \includegraphics[width=\textwidth]{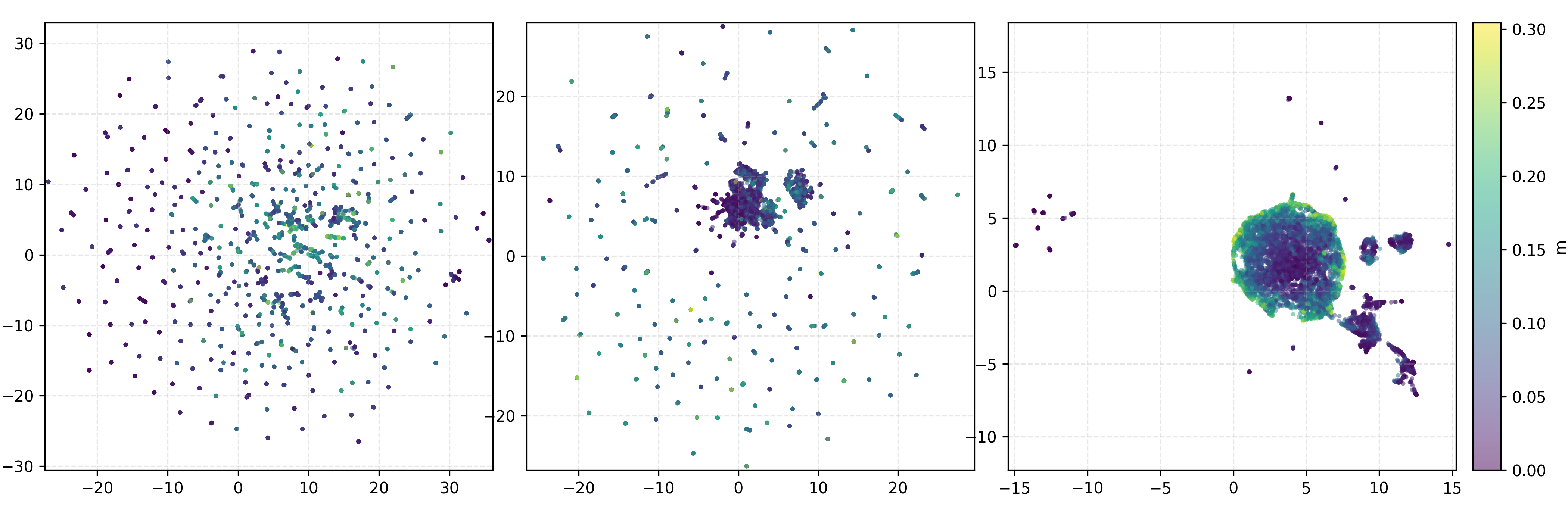}
  \caption{UMAP visualization colored by displacement magnitude. \textbf{Left}: LAPA; \textbf{Middle}: UniVLA; \textbf{Right}: Ours.}
  \label{fig:umap_mag}
\end{figure}

Figures~\ref{fig:umap_x}--\ref{fig:umap_mag} provide per-axis and
magnitude UMAP visualizations supplementing the codebook analysis in
\autoref{sec:exp_rq1}. Our codebook exhibits smooth, axis-aligned
gradients across all translational dimensions, whereas LAPA and UniVLA
show fragmented clusters with no consistent directional structure.

\paragraph{Additional target environments}
To assess adaptation beyond the five-task suite, we evaluate two
additional simulated target environments: one replaces the milk carton
with a round, slender ketchup bottle and updates the instruction;
the other shifts the camera azimuth by $5^{\circ}$.
Evaluated on 50 fixed initial states per environment, adaptation from
generated videos achieves $60\%$ and $74\%$ success, respectively,
rising to $94\%$ and $92\%$ after one round of self-evolution.
Here, self-evolution is performed separately per task, without
cross-task pooling.
These results are also available in our code repository.

\end{document}